\documentclass{Interspeech}
\usepackage{multirow}

\interspeechcameraready

\title{Mispronunciation Detection Without L2 Pronunciation Dataset in Low-Resource Setting: A Case Study in Finland Swedish}

\author[affiliation={1}]{Nhan}{Phan}
\author[affiliation={2}]{Mikko}{Kuronen}
\author[affiliation={2}]{Maria}{Kautonen}
\author[affiliation={2}]{Riikka}{Ullakonoja}
\author[affiliation={3}]{Anna}{von Zansen}
\author[affiliation={1}]{Yaroslav}{Getman}
\author[affiliation={1}]{Ekaterina}{Voskoboinik}
\author[affiliation={1}]{Tamás}{Grósz}
\author[affiliation={1}]{Mikko}{Kurimo}

\affiliation{Department of Information and Communications Engineering}{Aalto University}{Finland}
\affiliation{Department of Language and Communication Studies}{University of Jyväskylä}{Finland}
\affiliation{Department of Education}{University of Helsinki}{Finland}

\email{nhan.phan@aalto.fi}
\keywords{low-resource, Computer-Assisted Pronunciation Training
(CAPT), end-to-end, mispronunciation detection}

\usepackage{comment}
\usepackage{tipa}

\begin{document}

\maketitle

\begin{abstract}
    
    Mispronunciation detection (MD) models are the cornerstones of many language learning applications. Unfortunately, most systems are built for English and other major languages, while low-resourced language varieties, such as Finland Swedish (FS), lack such tools. In this paper, we introduce our MD model for FS, trained on 89 hours of first language (L1) speakers' spontaneous speech and tested on 33 minutes of L2 transcribed read-aloud speech.
    
    We trained a multilingual wav2vec 2.0 model with entropy regularization, followed by temperature scaling and top-k normalization after the inference to better adapt it for MD. The main novelty of our method lies in its simplicity, requiring minimal L2 data. The process is also language-independent, making it suitable for other low-resource languages. Our proposed algorithm allows us to balance Recall (43.2\%) and Precision (29.8\%), compared with the baseline model's Recall (77.5\%) and Precision (17.6\%).

    
    
\end{abstract}

\section{Introduction}

Pronunciation is a central learning goal for second language (L2) speakers, as it has a major impact on how well a speaker can be understood \cite{levis_pronunciation}. 
However, learning pronunciation in L2 is challenging, especially for adult learners, and therefore, applications that support learning by providing feedback to L2 speakers on their pronunciation are of great benefit to the learning process. Such applications are being researched and developed in the field of computer-assisted pronunciation training (CAPT), to which our study belongs.

Swedish is one of the two official languages of Finland, alongside Finnish. There are approximately 300,000 first language (L1) speakers of Swedish in Finland, and hundreds of thousands of Finns speak Swedish as L2 to varying degrees. The variety of Swedish spoken and most often studied in Finland is Finland Swedish (FS), a non-dominant variety of the pluricentric language Swedish. FS differs from Sweden Swedish (SweS) (i.e., varieties spoken in Sweden) mostly in pronunciation \cite{kautonen_finland_swedish_pronunc, ostman_standard_FS}, partly due to the influence of Finnish \cite{helgason_swedish_vs_FS}. Yet, there is no tool for practicing FS pronunciation to date. Such a tool is much needed, as the learning materials currently available online are almost exclusively based on SweS.

The main constraint in developing the MD model, not limited to FS but also other low-resource language varieties, is the scarcity of L2 pronunciation datasets \cite{lounis24_mdd_review}. We define L2 pronunciation dataset as an L2 speaker dataset containing, as a minimum requirement, phonetic level transcriptions or assessments done by qualified annotators for such task. Most notable examples of such datasets include L2-ARCTIC \cite{zhao_l2-arctic} or the much larger and detailed speechocean762 \cite{zhang_speechocean762}. Developing such corpora is a challenging task, as collecting L2 speech data is expensive and the annotation process is complex \cite{langlais98_swedish_md, oneil-etal-2024-developing_isiZulu}. There was an effort to develop an MD corpus for Swedish \cite{langlais98_swedish_md} decades ago, noting that the work was complex and time-consuming. To the best of our knowledge, Swedish still does not have a public L2 pronunciation corpus, and FS typically has fewer resources than SweS.

Instead of relying on an L2 pronunciation dataset - which would be far-fetched for low-resource language varieties like FS - we propose a framework for developing an MD model using only L1 spontaneous speech data. This approach is not an entirely novel idea. Lee and Glass \cite{lee15b_md_without_l2_train_data} previously introduced a similar concept by training MD models exclusively on L1 data. However, they used a proper L2 pronunciation corpus as the test set. Furthermore, they worked with the English language,  meaning their approach only has theoretical benefits for developing models for low-resource languages.

Previously, Phan et al. \cite{phan_captaina} have developed an MD model for Finnish without any L2 pronunciation data. However, this work relies on the fact that Finnish has very transparent (shallow) orthography, i.e., it is a language with close and consistent phoneme-to-grapheme correspondences. Finnish also has more resources (L1 data, speech research, etc.) than FS. Furthermore, FS lacks a standard pronunciation system (see \ref{sec:experiments}). Being inclusive and considering all dialects available in the dataset will lead to even more challenges in the low-resource setting.

Therefore, our research represents the first effort to develop an MD model for a low-resource language variety without relying on an L2 pronunciation dataset. We design a pipeline to develop the MD in that setting and introduce a simple algorithm to partly mitigate problems caused by the lack of L2 data. We also show that our FS model can detect some significant differences in pronunciation between FS and SweS. Our method is applicable to other low-resource languages facing similar challenges, as it requires minimal L2 data. This collaborative effort between technological and linguistic researchers addresses the practical needs of L2 learners of a low-resource language variety, promoting inclusive speech technology for all. The repository for our project is publicly available and open-source\footnote{\url{https://github.com/aalto-speech/FinSwedish/}}. 

\section{Dataset}

\begin{figure}[t]
  \centering
  \includegraphics[width=\linewidth]{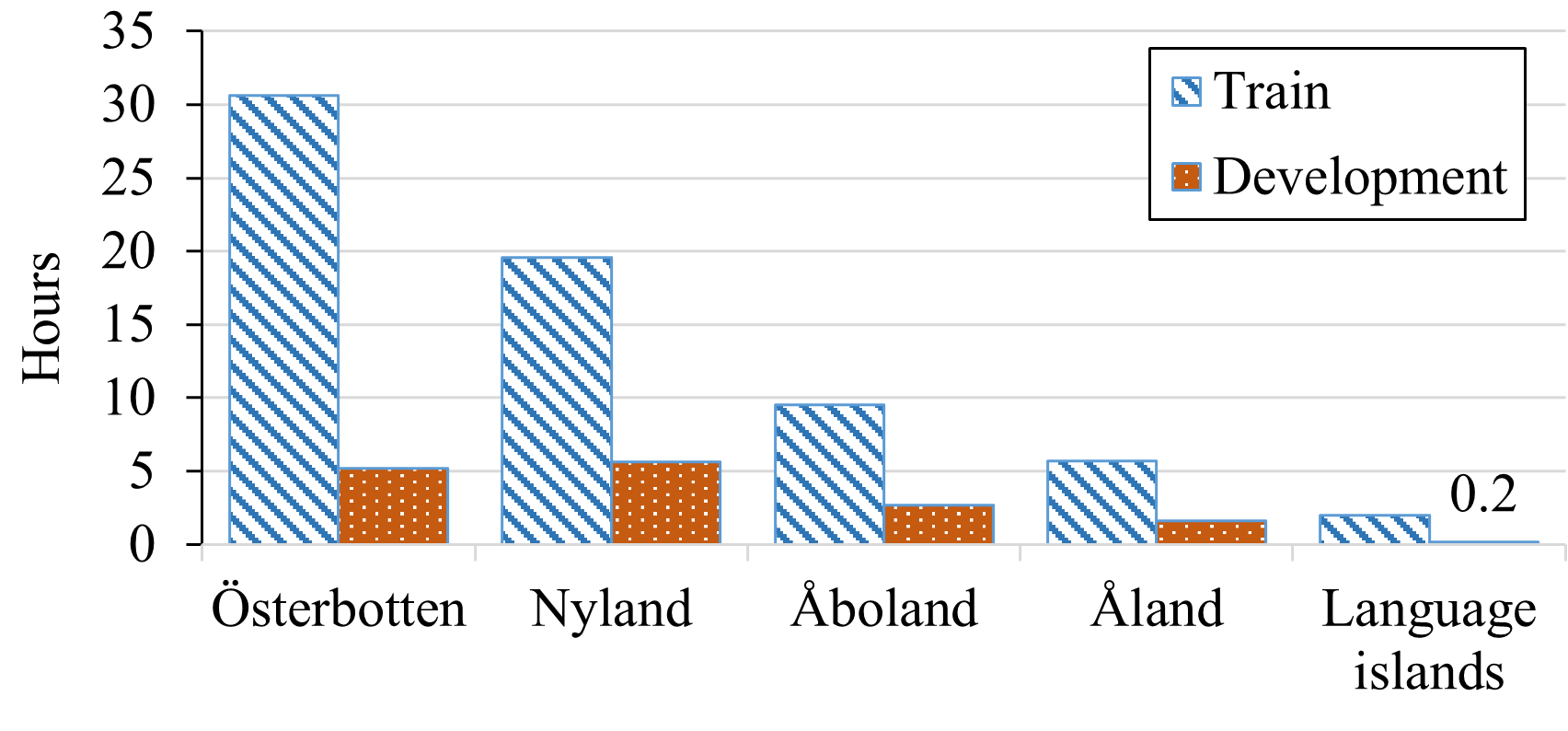}
  \caption{Distribution of regions in the Talko3 training and development sets. Most data is from Österbotten and Nyland.}
  \label{fig:talko3_split}
\end{figure}

Our FS data are derived from 3 corpora: the Talko3 spoken corpus\footnote{\url{https://www.sls.fi/talko}}, Aalto FS Parliament Automatic Speech Recognition (ASR) corpus \cite{raitolahti_fs}, and DigiTala \cite{al2023ASA_digitala}. The Talko3 is the largest FS dataset and it contains interviews where both speakers have FS as their L1. The speakers come from 5 regions where FS is mainly spoken in Finland: Österbotten, Nyland, Åboland, Åland, and language islands (small communities where Swedish is spoken in an environment dominated by a Finnish-speaking population). Most data is from Österbotten and Nyland, noting that the dialects spoken in Österbotten deviate most from the standard variety of FS. The dataset consists of both phonetic and orthographic transcriptions, with timestamps for each speaking segment. However, the duration of these audio samples varies, from less than 1 seconds to more than 2 minutes. While the total dataset consists of about 110 hours of recordings, for the purpose of training a model for the MD task, which typically only needs a maximum of 10 seconds, we have to filter out too short ("hm", "ah", etc.) and too long samples, as well as overlapping speech. In total, we used 83 hours of Talko3 corpus (see Figure \ref{fig:talko3_split}), with audio lengths of 2-25 seconds (mean 7 seconds). We randomly split the dataset by speaker into 82\% training (68 hours, 388 speakers) and 18\% development sets (15 hours, 80 speakers).

The Aalto FS Parliament ASR corpus has about 6 hours of transcribed speeches from the Finnish parliament between 2015 and 2020. The speakers are mainly from the Swedish People's Party of Finland, who are likely fluent in FS. Recording durations range between 0.5-15 seconds (mean 6 seconds). The processing pipeline ensures that no data is longer than 15 seconds, making this dataset more suitable for our purpose. We randomly split the dataset by speaker into 80\% training (5 hours, 64 speakers) and 20\% development sets (1 hour, 17 speakers). 

To test the MD model's performance, we use the read-aloud samples from the DigiTala corpus, where L2 speakers were asked to read aloud specific words or sentences. These include: \textit{banan} (banana), \textit{dyr} (expensive), \textit{kanske} (perhaps), \textit{jag tycker om att sjunga} (I like to sing), \textit{han kommer från kina} (he comes from China), \textit{vill du äta här?} (would you like to eat here?), \textit{på kvällen är jag på gymmet} (in the evening I am at the gym). Unlike the training data, the transcriptions in the DigiTala dataset are verbatim; the transcriber transcribed the speech orthographically according to what they heard. We leveraged this feature for MD purposes, as differences between the transcription and the target word would indicate mispronunciations noted by the transcriber. In total, we used 485 samples (about 33 minutes of speech data, with mean duration of 4 seconds) as a test set. While the DigiTala corpus is not specifically designed for MD, it can still be used as a test set for this purpose with some considerations:
\begin{itemize}
\item The transcriber was encouraged to note the mistakes made by the speakers but was not required. As a result, some mispronunciations may not be documented, especially during read-aloud tasks where the target speech is known beforehand.
\item The transcriber was not a phonetic expert and was not trained for the MD annotation task. 
\end{itemize}

Without a proper L2 pronunciation corpus, we cannot accurately evaluate the performance of the MD models. However, since all models evaluated on this data face the same limitations, we can use the relative performance of different models to select the most suitable one for our application. 

Although the DigiTala corpus has more data for spontaneous speech, we decided not to use this data for training or fine-tuning. Doing so would cause the models to overfit to the transcription style of the transcriber and overstate the results. We also did not use any SweS samples in the training, as previous research has shown that including SweS data does not significantly improve the models' understanding of FS~\cite{raitolahti_fs}.

\section{Framework and Experiment}

We define the correct identification of mispronunciation as true rejection (TR), the failure to detect mispronunciation as false acceptance (FA), and correct pronunciation that is misidentified as mispronunciation as false rejection (FR). For MD, we use Recall ($\frac{TR}{TR + FA}$), Precision ($\frac{TR}{TR + FR}$), and their harmonic mean F1 as our metrics~\cite{lounis24_mdd_review}.

\subsection{Principles}
\label{sec:principles}

Our FS MD models are developed based on three key principles: standard variety, intelligibility, and inclusive speech technology. Although FS contains many dialects, according to some estimates up to 80~\cite{ostman_standard_FS, ivars_dialects}, the variety most commonly used in teaching and studied by L2 speakers is the standard variety, which resembles the regional variety spoken in Central Nyland. Therefore, it is justified as a first principle that a technical application developed to support the learning of FS primarily aims at standard pronunciation 
while accepting some established and frequent pronunciation features in the FS dialects. 

The second principle guiding our research has been to support the learning of intelligible (i.e., communicatively functional) pronunciation, not native-likeness, as the latter has proven to be an unrealistic goal for most L2 speakers~\cite{levis_pronunc_target}. 

The third principle is inclusive speech technology for all. While we primarily aim for a standard pronunciation of FS, we include as many dialects as possible, even though that may reduce our model performance. For example, some dialects in Åland are closer to SweS than FS, and including them in training could lead to the MD model, in some cases, being unable to differentiate between FS and SweS. This is an expected result and, to some extent, satisfactory at the benefit of inclusively including all FS dialects. From the learner's point of view, the fact that the model accepts some SweS pronunciations is neither dangerous nor a major disadvantage, since these pronunciations are fully comprehensible to FS listeners.

To align our evaluation with these guiding principles, we focus on Precision over Recall. Recall prioritizes the detection of mispronunciation, while Precision focuses on detecting mispronunciation accurately. The social cost of Recall and Precision is not the same \cite{bachman_fundamental}. Considering our principles - emphasizing intelligibility and inclusivity - and the limitations of model performance in low-resource settings, we prioritize Precision over Recall. Aiming for higher Precision ensures that users receive feedback only on a few significant mispronunciations, which aligns with pedagogical recommendations \cite{eskenazi_feedback_education, ellis_corrective}.

\subsection{Experiments}
\label{sec:experiments}

While MD typically relies on a phonetic system \cite{lounis24_mdd_review}, we rely instead on the orthographic version of the Talko3 dataset for two reasons. Firstly, the phonetic annotation is expensive (the Talko3 corpus took over 10 years and 20 people for the annotation work) and, therefore, often unavailable for L2 speech in low-resource languages. In the case of FS, we do not have any L2 pronunciation dataset. Also, as we lack sufficient training data (even for L1), adopting the orthography model allows us to leverage data from other ASR speech corpora. Such data may be available in the future.

Secondly, many words in FS have several possible pronunciations. For example, \textit{fara} and \textit{göra} can be pronounced with a short first vowel \textipa{[fara]} and \textipa{[j\oe ra]} or a long vowel \textipa{[fA:ra]} and \textipa{[j\oe:ra]} like in SweS. Ideally, the MD model would accept both short and long first vowels as correct pronunciation. A model trained on the orthographic form could accommodate these variations. This satisfies our research principles mentioned in \ref{sec:principles}.

Models trained exclusively on L1 data may not perform well with out-of-domain L2 speech. To mitigate this issue, we leveraged wav2vec 2.0 architecture \cite{baevski2020wav2vec} and the XLS-R pre-trained model \cite{babu_xlsr} with 300 million parameters \footnote{https://huggingface.co/facebook/wav2vec2-xls-r-300m}. XLS-R models were pre-trained with 436,000 hours of unlabeled speech from 128 languages and can improve the MD performance for L2 speakers \cite{phan_captaina, peng_mdd_wav2vec2}. As explained above, the orthography models will predict alphabet characters rather than phonemes. To get the character level score, we used the forced alignment algorithm \cite{kurzinger_fa_ctc} on the Connectionist Temporal Classification (CTC) \cite{graves_ctc} outputs of the wav2vec 2.0 models.

Furthermore, we also implemented the maximum entropy regularization \cite{liu_entropy_ctc} during training with hyperparameter $\beta\%$. In addition to improving the model's robustness with L2 speech, higher $\beta\%$  also helps reduce the peakiness of the CTC output, allowing for a better scoring of the MD models~\cite{phan_captaina}.

\subsection{Temperature scaling and top-k normalization}

MD models typically rely on a threshold level, $\theta$, to determine a mispronunciation \cite{lounis24_mdd_review, phan_captaina, xu_explore_wav2vec_md} based on the probability output, similar to the goodness of pronunciation score \cite{witt_gop}. In our case, we do not have any L2 data for fine-tuning, so even with entropy regularization, the CTC output tends to be overconfident. This overconfidence emphasizes some limitations of the threshold method. Firstly, we lack a validation set to determine the optimal $\theta$ value. Furthermore, adjusting  $\theta$ merely represents a trade-off between recall and precision; lowering the threshold can cause many mispronunciations to be ignored. Additionally, the results are binary, categorizing pronunciation as either correct or incorrect.


Instead, we set $\theta$=50\% (the simplest probability-based threshold) and proposed a simplified version of the temperature scaling to calibrate the probability output of our models. Temperature scaling with a single scalar parameter $T$ is the simplest, fastest, and often the most effective method for probabilistic calibration \cite{guo_temperature_scale}. As temperature scaling requires validation, we propose an even simpler algorithm to circumvent this requirement, called temperature scaling and top-k normalization. It simply softens the output probability by dividing the logit vector by a temperature hyperparameter $T$. We then rescale the top-k probability to get the score in the range [0, 1]. The algorithm applied to the logit outputs after inference is as follows:

\begin{enumerate}
\item Divide the logit vector $z$ by $T$. The predicted labels' probabilities, $P$, are obtained by softmax function $\sigma$: $P= \sigma(z/T)$
\item Determine the set $\text{top-k}$ of labels whose probabilities are the largest $k$ in $P$. Divide $P$ by the scaled probability of $\text{top-1}$. $\text{top-1}$ will be normalized to a score of 1, and $\text{top-k}$ will be rescaled to higher values: $\text{top-k}_{\text{score}} = \frac{P}{\text{top-}1_{\text{P}}}$
\item Replace the probabilities of $\text{top-k}$ with the normalized $\text{top-k}_{\text{score}}$, keep the original probabilities of the other labels.
\end{enumerate}

Considering a simple CTC output with four labels and their probabilities: L (0.998), R (1e-3), K (1e-5), and padding token PAD (9.9e-4). This is a typical CTC output, showing overconfidence in L and under-estimate the probability of R. By applying the above algorithm with $T$=10, $k$=3 to the logit outputs, we obtain the scores: L (1), R (0.5), K (1e-5), PAD (0.5). Temperature scaling reduces the confidence level in the predicted label by adjusting the probability distribution. As a result, it reduces the gap between the predicted label's score and the scores of other labels, even if their initial probabilities are much smaller. This allows us to accept both L and R as correct. Moreover, since the scores are normalized to the range [0, 1], it allows for finer grading beyond a binary decision, for instance, considering setting an arbitrary range of 50\%-75\% as ``partially correct". 

\begin{figure}[th]
  \centering
  \includegraphics[width=\linewidth]{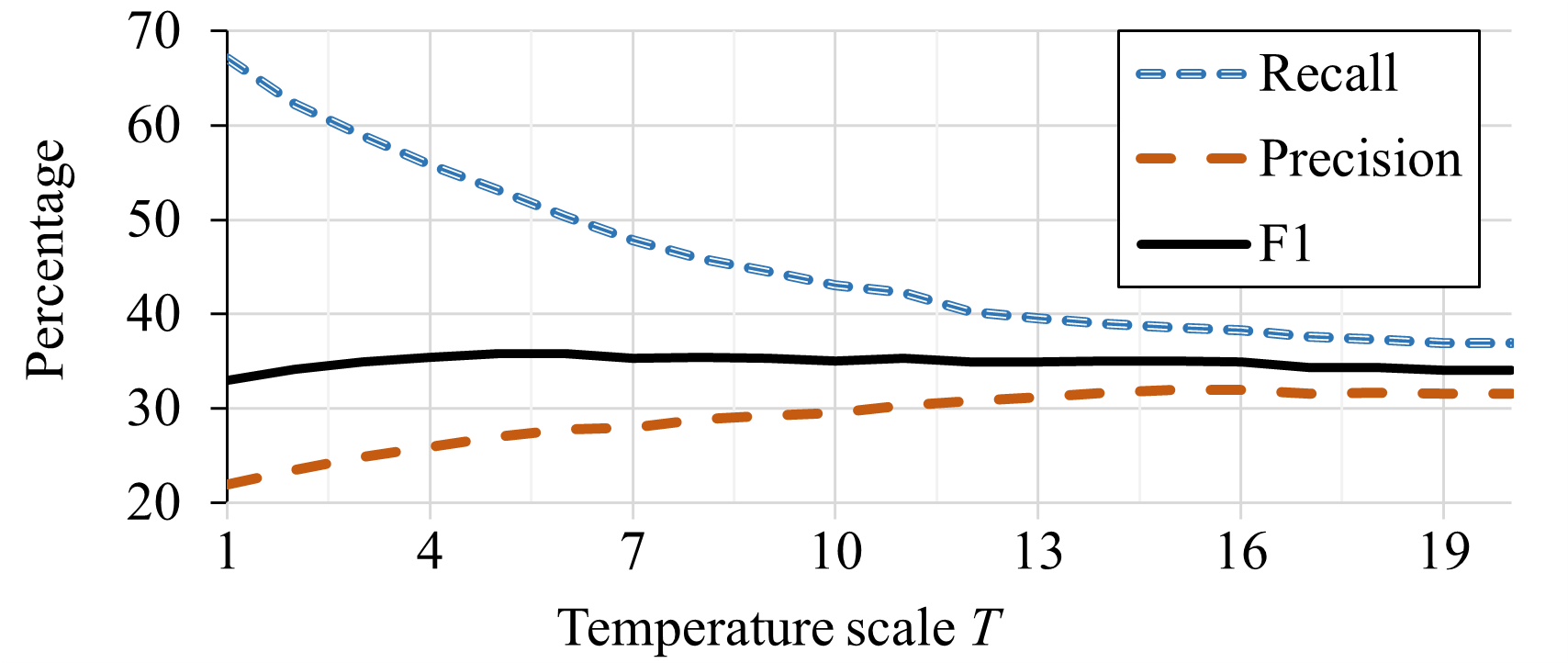}
  \caption{XLS-R model performance per $T$ value, with entropy $\beta$=20\% and top-k with $k$=3. As T increased larger than 10, performance metrics did not change significantly.}
  \label{fig:temperature_scaling}
\end{figure}

We do not have to optimize for $T$; as long as $T$ is large enough to compensate for the peakiness of CTC output, the choice of $T$ does not significantly affect model performance (see Figure \ref{fig:temperature_scaling} and Section \ref{sec:results} for further discussion on changing $T$). As we do not have any validation set, we choose $k$=3, so the normalization is primarily applied to another character, as the padding token is usually among the top-3. In our analysis of the development set, whenever a non-padding label has the highest probability, the padding token is in the top-3 99\% of the time. This allows us to maintain Recall without optimizing $T$. The algorithm is designed to favor Precision, aligning with our principle mentioned in \ref{sec:principles}.

\section{Results}
\label{sec:results}
Our results are shown in Table \ref{tab:results}. The baseline model ($\beta$=0\%, $T$=0) has very low Precision, which is expected given the diversity of FS dialects and the limited amount of data available to adequately capture this variability. This limitation is evidenced by Otto-Ville \cite{raitolahti_fs} in his Talko3 region-wise L1 ASR performance analysis, with word error rate varying from 27.6\% to 35.6\%. With the same $T$ value, entropy regularization helps increase both Precision and Recall, similar to the results reported by Phan et al., \cite{phan_captaina}. However, in our case, the $\beta\%$ needs to be higher for a significant difference.

\begin{table}[t]
  \caption{Models performance at character-level and word-level. Due to the lack of an L2 pronunciation corpus, the results do not reflect the actual performance of those models. All temperature scaling has top-k with $k$=3}
  \label{tab:results}
  \centering
  \begin{tabular}{ c c c c c c }
    \toprule
    & \textbf{Entropy $\beta$} & \textbf{$T$} & \textbf{Precision}  &  \textbf{Recall}  & \textbf{F1}  \\
    \midrule    
    \multirow{3}{2.5em}{Character level}& 0\% & 0 & 17.6\% & \textbf{77.5\%}& 28.7\% \\
    & 0\% & 10 & 29.4\% & 37.3\%& 32.8\% \\    
    & 20\% & 10 & \textbf{29.5\%} & 43.2\%& \textbf{35.0\%} \\
    \midrule    
    \multirow{3}{2.5em}{Word level}& 0\% & 0 & 31.3\% & \textbf{85.0\%}& 45.8\% \\
    & 0\% & 10 & 41.9\% & 49.5\%& 45.4\% \\    
    & 20\% & 10 & \textbf{42.4\%} & 56.1\%& \textbf{48.3\%} \\
    \bottomrule
  \end{tabular}
  
\end{table}

The most significant improvement comes from the temperature scaling and top-k normalization, allowing us to balance Precision and Recall and consequently increase F1. It should be emphasized that, from our perspective, there is no single best model. Since we do not have the validation data, we do not think optimizing for the best model is necessary for our case. We select the XLS-R pre-trained model with 20\% entropy, $T$=10, $k$=3 (20\%-10-3) as our primary model for the analysis. In practice, any XLS-R-20\% can be used, as the temperature scale $T$ and top-k are applied after inference, with almost no computing cost. We propose changing the $T$ value dynamically to meet the learners' needs, i.e., increasing $T$ if they cannot pronounce correctly to encourage them to practice more or lowering $T$ if they want advanced pronunciation practice. We believe this dynamic approach can also be helpful for individuals with speech impairment. As we can see from Figure \ref{fig:temperature_scaling}, the algorithm maintains a balance between Recall and Precision, regardless of $T$.

Since we do not have an L2 pronunciation dataset, we rely on the orthographic transcription of all corpora. As a result, the MD results are at the character level instead of the phoneme level. As the model has no mapping between grapheme and phoneme, the feedback at the character level is not reliable. Therefore, while we develop our models at the character level, we use word-level in our MD application and also report the word-level metrics. We agree with Lee and Glass \cite{lee15b_md_without_l2_train_data} that word-level feedback is valuable for beginner language learners while maintaining a reliable level of performance.

Finally, to validate and analyze the capabilities of the 20\%-10-3 model, two phonetics experts recorded 34 sentences - 16 with at least one mispronunciation at word-level and 18 with none. On this small test set, the model achieved 33\% Recall and 80\% Precision at the word-level. As high Precision is preferred for our CAPT application, these preliminary results are encouraging. However, due to the limited sample size, we cannot draw definitive conclusions about the model’s performance.

\subsection{FS and SweS pronunciation}

Using the 20\%-10-3 model and the recordings from Mozilla's Common Voice 17.0 Swedish (CV Swedish) dataset (train, development, and test set) \cite{commonvoice}, we evaluated a few selected words (details can be checked in our repository) that have significant differences in pronunciation between FS and SweS. We compared the percentage of mispronunciation detection with the average words of the same length. We hypothesize that if our model can differentiate between FS and SweS, those words would have much higher detected \% (mispronunciation) than the average \%. The results are in Table \ref{tab:fs_and_swedish}. 

\begin{table}[t]
  \caption{The ability of the 20\%-10-3 model to detect the differences between FS and SweS. The Average \% is the Detected \% of other words with the same length in the CV Swedish. $p$ value is from the one-side proportional z test.}
  \label{tab:fs_and_swedish}
  \centering
  \begin{tabular}{ c c c c c }
    \toprule
    \textbf{Group} & \textbf{Count} & \textbf{Detected \%} & \textbf{Average \%}  &  \textbf{$p$} \\
    \midrule                
    \textit{fara, göra} & 230 & 5.2\%& 12.0\% & 8e-4    \\    
    \textipa{/rt/}, \textipa{/u:/} & 172 & 6.4\%& 15.8\% & 3e-4    \\    
    Others     & 187 &31.6\%& 14.1\% & 4e-12   \\    
    \bottomrule
  \end{tabular}  
\end{table}

Our model detected less mispronunciation of \textit{fara, göra} than average, which indicates the model tends to consider SweS pronunciation of \textit{fara, göra} as correct pronunciation. This aligns with our objectives. However, in some cases involving \textipa{/rt/} and \textipa{/u:/}, such as \textit{bort} (FS \textipa{[bort:]} vs. SweS \textipa{[bo\:t:]}) and \textit{telefon} (FS \textipa{[tElEfu:n]} vs. SweS \textipa{[tElEfo:n]}), the model also accepts most SweS pronunciations, which is not our desired outcome. 

For some other words, notably \textit{sju} (FS \textipa{[S0:]} vs. SweS \textipa{[\texththeng 0:]}), \textit{skjorta} (FS \textipa{[Su:rt:a]} vs. SweS \textipa{[\texththeng u\:t:a]}), words beginning with ``dj" like \textit{djur} (FS \textipa{[\t{dZ}0:r]} vs. SweS \textipa{[J0:r]}), and \textit{tjejer} (FS \textipa{[\t{tS}Ej:Er]} vs. SweS \textipa{[CEj:Er]}), the model detects 31.6\% of the samples as a mispronunciation, which is statistically higher than the average 14.1\%. This indicates that, in these cases, our FS model can detect the differences from SweS pronunciation.

\subsection{Limitations}

Our work comes with several challenges inherent to working in low-resource settings. The training data is regionally imbalanced (Figure \ref{fig:talko3_split}), which could bias our MD model toward particular dialects. Furthermore, the limited training data prevent us from achieving a balanced dataset across other variables, such as gender and age. Addressing these potential biases requires further research to fully understand and mitigate them.

The absence of an L2 pronunciation corpus poses a challenge for accurately measuring the performance of our model. However, in a low-resource language context, creating a phonetically annotated L2 dataset is often impractical due to constraints in time, funding, and available expertise. Pursuing such a dataset would be misaligned with the primary objectives of this study, which emphasize practical utility in low-resource settings. Our goal is to support intelligible L2 pronunciation in practical CAPT applications, and we believe our approach remains important in real-world scenarios. Our results provide valuable insights into possible approaches to mispronunciation detection in resource-constrained settings.
 
\section{Conclusion}

In this paper, we detailed our pipeline and algorithm for developing MD models for a low-resource dialect. We relied entirely on L1 data and tested on a minimal L2 ASR corpus. Our model was able to detect some significant differences between FS and SweS while recognizing the variations within FS dialects. However, there were cases where the model failed to identify the differences. Further research is needed to identify potential issues and improve the model's performance.

Our key contribution is the introduction of the simplified temperature scaling and top-k normalization algorithm, which allows us to reach a balanced and reasonable level of performance even with minimal data. We believe our findings can help in the development of similar applications for under-researched languages or language varieties, thus promoting inclusivity in speech technology.

\section{Acknowledgements}
We would like to thank the following projects and funding agencies: NordForsk through the funding to ``Technology-enhanced foreign and second-language learning of Nordic languages'' (project number 103893); Research Council of Finland through the funding to ``Digital support for training and assessing second language speaking'' (grant no 322619, 322625, 322965), and ``Automatic assessment of spoken interaction in second language'' (grant no 355586, 355587, 355588). The computational resources were provided by Aalto ScienceIT.

The development of the mobile app was funded by the Kielibuusti project in 2023. We are grateful to Apollo Ailus and Kia Raitanen for their contributions to user research and engagement, and to Aalo Kailu, who designed the original user interface of the app.

\bibliographystyle{IEEEtran}
\bibliography{mybib}

\end{document}